\documentclass[conference]{IEEEtran}
\IEEEoverridecommandlockouts

\usepackage{tikz}
\usepackage{textcomp}
\usepackage{hyperref}
\usepackage{lipsum}

\newcommand\copyrighttext{%
  \footnotesize \textcopyright 2025 IEEE. Personal use of this material is permitted.  Permission from IEEE must be obtained for all other uses, in any current or future media, including reprinting/republishing this material for advertising or promotional purposes, creating new collective works, for resale or redistribution to servers or lists, or reuse of any copyrighted component of this work in other works.}
\newcommand\copyrightnotice{%
\begin{tikzpicture}[remember picture,overlay]
\node[anchor=south,yshift=10pt] at (current page.south) {\fbox{\parbox{\dimexpr\textwidth-\fboxsep-\fboxrule\relax}{\copyrighttext}}};
\end{tikzpicture}%
}

\usepackage{cite}
\usepackage{amsmath,amssymb,amsfonts}
\usepackage{algorithmic}
\usepackage{graphicx}
\usepackage{textcomp}
\usepackage{xcolor}
\usepackage{lipsum}
\usepackage{float}
\usepackage{diagbox}
\usepackage{balance}
\usepackage{array}
\usepackage{footnote}
\usepackage{booktabs}
\usepackage{adjustbox}
\usepackage{multirow}
\usepackage{tabularx}
\usepackage[per-mode=symbol]{siunitx}
\def\BibTeX{{\rm B\kern-.05em{\sc i\kern-.025em b}\kern-.08em
    T\kern-.1667em\lower.7ex\hbox{E}\kern-.125emX}}
\begin{document}

\title{Efficient Continual Learning in Keyword Spotting using Binary Neural Networks}



\author{
    \IEEEauthorblockN{
        Quynh Nguyen-Phuong Vu\IEEEauthorrefmark{1},
        Luciano Sebastian Martinez-Rau\IEEEauthorrefmark{1},
        Yuxuan Zhang\IEEEauthorrefmark{1},
        Nho-Duc Tran\IEEEauthorrefmark{1}, \\
        Bengt Oelmann\IEEEauthorrefmark{1},
        Michele Magno\IEEEauthorrefmark{2} and
        Sebastian Bader\IEEEauthorrefmark{1}
    }
    \IEEEauthorblockA{
        \IEEEauthorrefmark{1}\textit{Department of Computer and Electrical Engineering, Mid Sweden University}, Sundsvall, Sweden \\
        \IEEEauthorrefmark{2}\textit{Department of Information Technology and Electrical Engineering, ETH Zürich,  Zürich, Switzerland} \\
        Email:
        quynh.nguyenphuongvu@miun.se
    }
}

\maketitle
\copyrightnotice

\begin{abstract}
Keyword spotting (KWS) is an essential function that enables interaction with ubiquitous smart devices. However, in resource-limited devices, KWS models are often static and can thus not adapt to new scenarios, such as added keywords. To overcome this problem, we propose a Continual Learning (CL) approach for KWS built on Binary Neural Networks (BNNs). The framework leverages the reduced computation and memory requirements of BNNs while incorporating techniques that enable the seamless integration of new keywords over time. This study evaluates seven CL techniques on a 16-class use case, reporting an accuracy exceeding 95\% for a single additional keyword and up to 86\% for four additional classes. Sensitivity to the amount of training samples in the CL phase, and differences in computational complexities are being evaluated. These evaluations demonstrate that batch-based algorithms are more sensitive to the CL dataset size, and that differences between the computational complexities are insignificant.
These findings highlight the potential of developing an effective and computationally efficient technique for continuously integrating new keywords in KWS applications that is compatible with resource-constrained devices.

\end{abstract}


\section{Introduction}

%


Keyword spotting (KWS) has drawn much interest over the past decade due to its widespread implementation in modern speech recognition systems, enabling hands-free interactions with devices through voice commands. Noticeable examples include the wake-up commands 'Ok, Google' or 'Hey, Siri', which invoke smartphone voice interactions, and the activation phrase 'Alexa' in Amazon's smart home devices \cite{google, alexa}. Additionally, keyword-based triggers are now commonly used to enable driver assistance features in modern vehicles \cite{kws_veh,7225681} or applied to smart home assistants \cite{smarthome, homerasp}. In the near future, it is expected that KWS will find its way into more application scenarios, including applications on resource-constrained devices, such as wearables and low-cost sensors.

To tackle such application scenarios, recent research has investigated KWS models that are compatible with resource-constrained computational platforms, such as microcontroller units (MCUs). Zhang et al. \cite{Zhang2017HelloEK} proposed a quantized 8-bit fixed-point Convolutional Neural Network (CNN) model deployed on an MCU, demonstrating no loss in accuracy due to the model compression. Another model, known as LiCoNet, has been introduced in \cite{Lico}, leveraging int8 linear operators during inference to improve computational efficiency while maintaining classification performance. In order to further compress model size, Cerutti et al. investigated Binary Neural Networks (BNNs) for KWS applications \cite{submW}. They compared models with binary weights, as well as binary weights and binary activations, with the equivalent full-precision model, demonstrating negligible loss in accuracy but significant implementation benefits on a RISC-V MCU. These solutions enable cost-efficient, low-power, real-time voice processing suitable for battery-operated, embedded, and IoT applications in cost-sensitive markets. However, they are limited to static models that do not change after their initial deployment.

In this paper, we investigate the possibility to adjust KWS models for resource-constrained devices after their deployment. Specifically, we apply Continual Learning (CL) algorithms to a BNN-based KWS model in order to include new keywords. Several CL algorithms have been proposed in previous research, including \cite{9746488, Michieli2023OnlineCL, dual}. Huang et al. introduced a CL algorithm for small-footprint KWS, consisting of a network instantiator and a shared memory connected to a set of classification sub-networks \cite{9746488}. In \cite{Michieli2023OnlineCL}, the authors proposed statistical pooling to extract enriched temporal information from speech features and a Gaussian-based classifier to model class representations on the enriched space with a shared covariance matrix. Yang et al. proposed a replay-based CL algorithm for KWS that employs a dual-memory multi-modal structure, using short-term and long-term models to adaptively learn new and past knowledge while leveraging a class-balanced selection strategy based on confidence scores to enhance generalization and robustness \cite{dual}. However, these methods lead to considerable model complexity and higher memory consumption due to the storage of training data to effectively adapt to new keywords, which is not ideal for resource-constrained MCUs.

We propose a dynamic KWS approach combining resource-efficient CL algorithms and a BNN model. This approach leverages the advantages of binarized models in terms of computational and memory efficiency, while incorporating techniques that enable the seamless integration of new keywords over time. 

Our main contributions are as follows:
\begin{itemize}
    \item We implement and evaluate seven state-of-the-art CL algorithms on a BNN model optimized for KWS on resource-constrained MCUs. The performance comparison includes classification accuracy and computational complexity in terms of the number of Floating Point Operations (FLOPs) during backpropagation. By integrating these methods, our goal is to develop a scalable and energy-efficient KWS approach that can continuously learn new keywords while retaining previously acquired knowledge.
   
    \item We examine the impact of introducing varying numbers of new classes in KWS systems, assessing their ability to generalize to newly added keywords while maintaining performance on existing ones. These findings contribute to the development of more robust and scalable KWS models suitable for deployment on resource-constrained devices like MCUs.

    \item We investigate the effect of the number of training samples during CL, evaluating the trade-off between classification accuracy and the requirement of new training data. These findings guide practical implementations by identifying the algorithms sensitivity to training set size.
\end{itemize}

The rest of the paper is organized as follows: Section 2 reviews related work and mentions the limitations of existing approaches. Section 3 provides a description of the dataset and outlines the methodology used in our study. Section 4 presents the experimental results, while Section 5 summarizes the conclusions and explores potential embedded platforms of the proposed approach.

\section{Background}

\subsection{Binary Neural Network}

KWS models are commonly deployed on mobile edge devices, where data is processed locally to maintain low energy consumption and address user privacy concerns associated with transmitting data to the cloud \cite{google}. However, neural networks are computationally intensive and require significant memory, which poses challenges for deploying them on low-power embedded systems \cite{deepcomp}. To overcome this, various techniques have been developed to reduce their computational complexity. An effective method is model compression through quantization. 
Quantization reduces numerical precision to save memory and data bandwidth, decreasing storage and computation during inference \cite{extremelow}. 
The most extreme form of quantization is binarization, reducing the precision of both weights and neuron activations to a single bit \cite{binaryNN, XNOR}. Several resource-constrained systems have utilized BNNs for KWS, demonstrating remarkable results \cite{510-nW, 22nm}. In \cite{submW}, for example, a 71-fold reduction in overall energy cost was achieved, while the accuracy dropped only by 2\%. However, to the best of our knowledge, the advantages of BNNs have not been exploited in CL scenarios for KWS applications. 

\subsection{Continual Learning}

\begin{figure}[t]
\centering
      \includegraphics[width=\columnwidth]{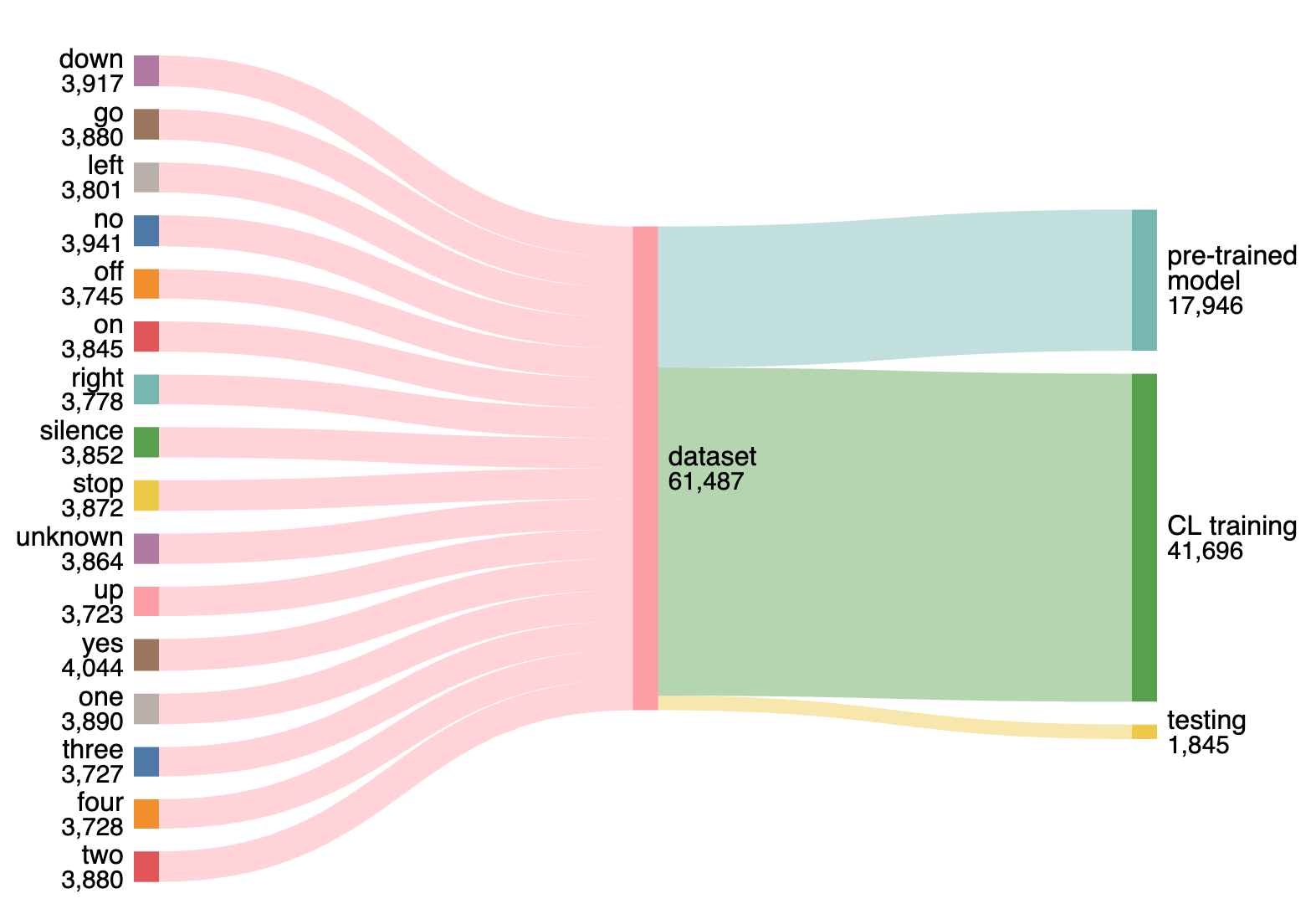}
      \caption{Dataset composition and separation based on the 16 selected classes of the Google Speech Commands V2 dataset.}
      \label{fig_sankey}
\end{figure}

\begin{figure*}[t]
    \centering
      \includegraphics[width=0.85\textwidth]{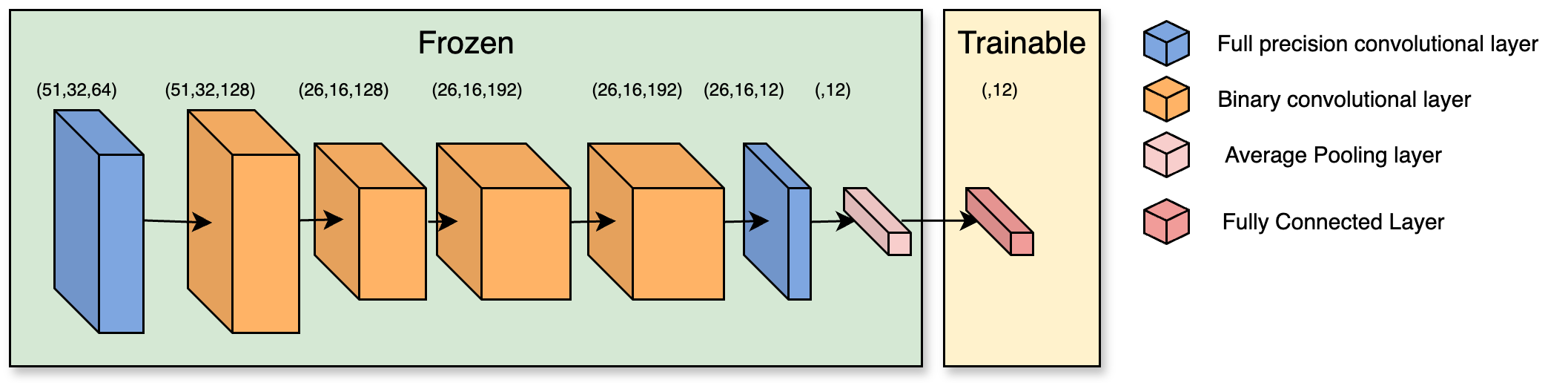}
      \caption{Overview of the BNN model: The first and last convolutional layers are in full precision while the other convolutional layers are binarized. A final fully-connected layer is added to be trainable, whereas the remaining model remains frozen.}
      \label{fig_BNN}
\end{figure*}

CL describes the setting in which a ML model is continuously trained from a sequence of data individually, or in small batches, while maintaining similar performance levels to models trained on all data simultaneously \cite{10.1109/TPAMI.2024.3367329}. Applications of CL may include the provision of new classes, additional samples of previously learned classes, changing environmental conditions, and diverse contexts.

CL scenarios can be broadly categorized as follows:
\begin{enumerate}
    \item \emph{New instances}: the model encounters new data of previously learned classes, requiring it to consolidate and expand its existing knowledge.
    \item \emph{New classes}: the model is introduced to entirely new classes, necessitating the integration of new concepts without compromising prior learning.
    \item \emph{New instances and classes}: the model faces both new instances of known classes and entirely new classes \cite{FRENCH1999128}. 
\end{enumerate}

We adopt the \emph{new instances and classes} scenario in our research as it provides a comprehensive and realistic learning environment. 

Initial CL algorithms have utilized Regularization-based methods \cite{ewc, si, Michieli2023OnlineCL}, Rehearsal methods \cite{gem}, and Generative replay methods \cite{ icarl}. However, these methods involve complex learning processes and require significant data storage, making them less suitable for deployment on resource-constrained devices.


To overcome this, resource-efficient CL algorithms have been proposed recently, making them feasible for operating on resource-constrained devices. A common approach in these algorithms is to only adjust parameters in the last model layer \cite{avi2022incremental}. The algorithms include:
\begin{itemize}
    \item \emph{TinyOL}, which computes the prediction error after one inference and adjusts the weights and biases through stochastic gradient descend. A variant exists, \emph{TinyOL with batches}, which applies backpropagation after the inference of a batch of samples.
    \item \emph{TinyOL v2}, which operates similarly to \emph{TinyOL}, but updates the parameters (weight and bias) of new classes, without modifying the parameters of the initial classes. TinyOL v2 can be operated based on a single sample or a batch of samples (\emph{TinyOL v2 with batches}).
    \item \emph{Learning Without Forgetting (LwF)}, which uses two parallel last classification layers: a training layer, which keeps learning and updating with new data, and a copy layer, which remains unchanged. When making predictions, the model applies the loss function on both layers to ensures that learning the new task does not erase what was learned before. 
\emph{LwF with batches} LwF with batch follows the same approach but does not keep the copy layer fixed throughout the entire training process. Instead, it updates the copy layer’s values after each batch is completed.

    \item \emph{Copy Weight with Reinit (CWR)}, which leverages two classification layers with a weighted backpropagation approach: the training layer, is updated continuously at every training step, while the  consolidated layer, is updated only after each batch is completed. During training, predictions are made using only the training layer, and its weights and biases are adjusted using the standard TinyOL method. This approach helps the model learn efficiently while balancing updates between short-term and long-term knowledge

\end{itemize}


\section{Materials and Methods}
\label{s3}

In this section, we describe the dataset and the BNN model utilized in the study, as well as the training process, including initial pre-training and continuous training. 

\subsection{Dataset}

\begin{figure}[t]
    \centering
      \includegraphics[width=\columnwidth]{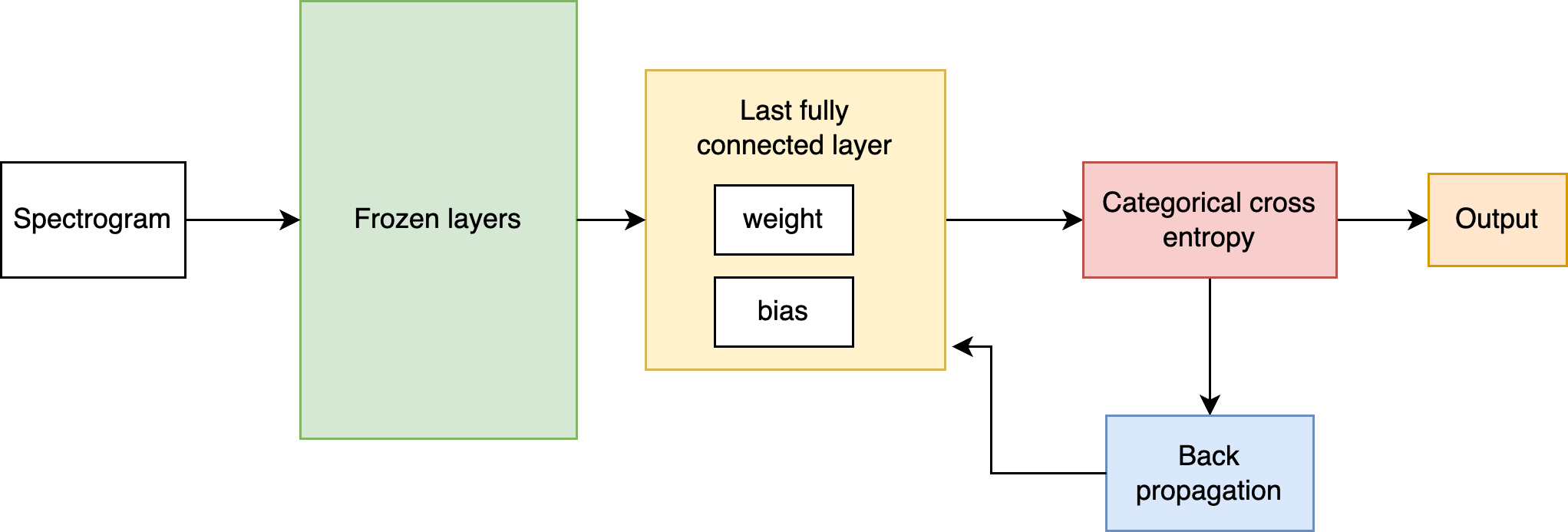}
      \caption{Diagram of the CL approach: Weights and biases of the last fully connected layer are updated through backpropagation based on categorical cross entropy loss.}
      \label{fig_backpropa}
\end{figure}

In this study, the Google Speech Commands V2 dataset is used, which contains 105,000 speech samples spanning 35 different keywords \cite{GSC}. Each speech sample has a duration of one second and is sampled at 16 kHz. In accordance with previous studies, we use a total of 16 classes, including the command keywords: “Yes”, “No”, “Up”, “Down”, “Left”, “Right”, “On”, “Off”, “Stop”, “Go”; the numeric keywords: "One", "Two", "Three", "Four"; as well as the two classes “silence” (i.e., no word spoken), which consists of one-second clips of random background noise, and an “unknown” word, which utilized the remaining 21 keywords. The selection of the 16 classes considers complexity, includes essential voice commands for smart devices, and agrees with previous studies allowing for benchmarking of the obtained results. In total, the selected dataset contains 61487 samples. 



Figure \ref{fig_sankey} illustrates the dataset division into subsets for initial training, CL, and test. Three percent of the data (i.e., 1,845 samples) are used as a test set to evaluate the algorithms. The remaining 97\% have been divided in two subsets to train the BNN model. The first subset is used to generate the pre-trained KWS model, and consists of 12 classes, excluding the four numeric keywords. The second subset contains samples of all 16 classes, and is used for the CL task. Initial experiments show that pre-training the BNN model with 40\% of the available data is sufficient to generate an accurate model. This approach achieves a 91.14\% accuracy, which is only a 0.5\% reduction compared to using the full dataset. However, the model is limited to classifying 12 out of 16 classes. The remaining samples are used for CL.

In order to preprocess the speech samples, the original time-domain audio clips are converted to the Log-mel frequency domain using 25 ms windows, a 10 ms hop size, and 64 Mel filters spanning 50–7500 Hz to reduce data dimensionality. This signal preprocessing step can also be executed using an external low-power analog front-end, specifically designed for embedded systems, which provides an efficient implementation for practical applications \cite{AFE}.
\begin{figure}[t]
    \centering
    \includegraphics[width=.95\columnwidth]{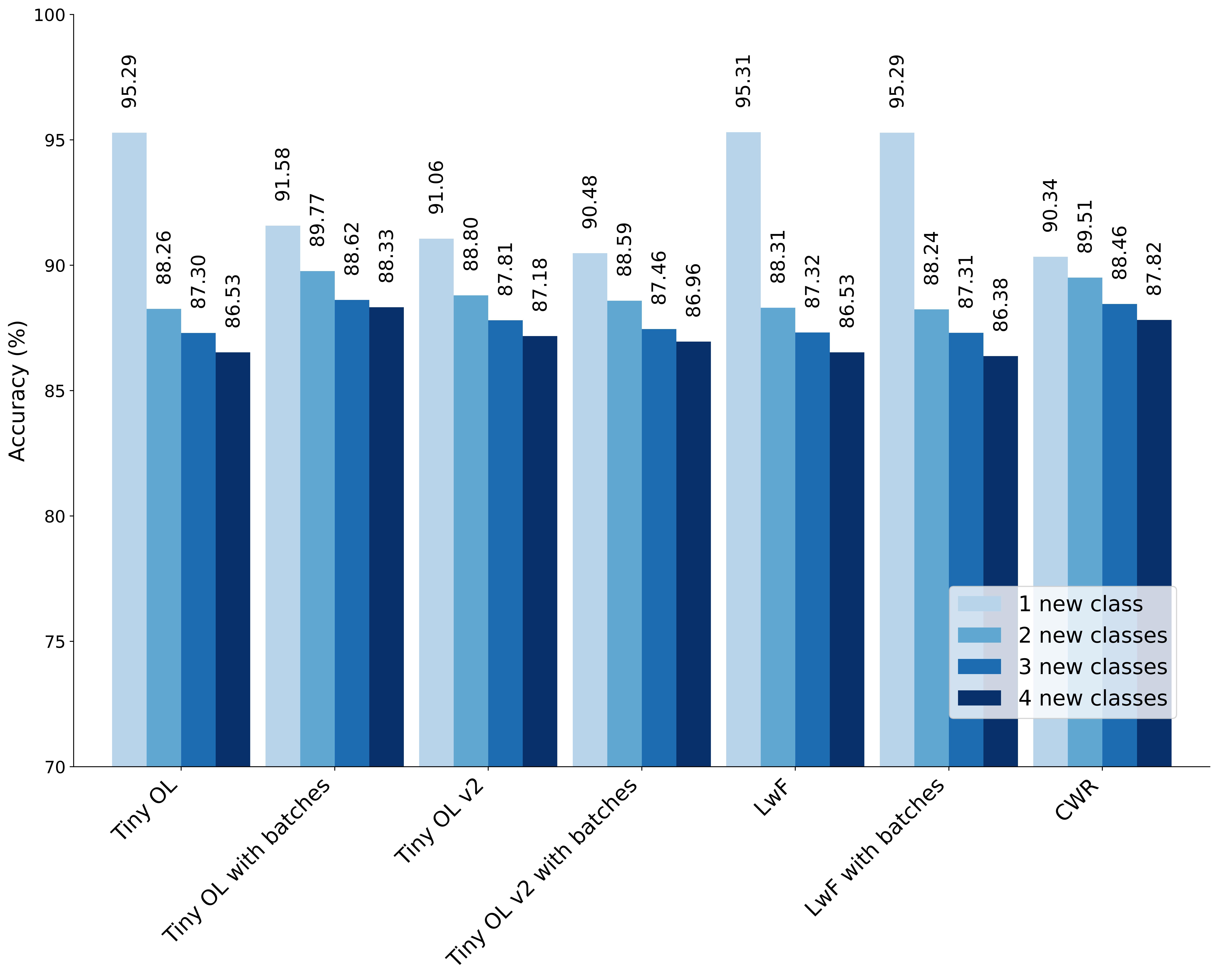}
    \caption{Accuracy of the original 12 classes across the seven CL algorithms and different number of new classes.}
    \label{fig_12}
\end{figure}

\subsection{BNN model}

The BNN model in our study is based on the architecture proposed in \cite{submW}, which is depicted in Fig. ~\ref{fig_BNN}. The binarized version of this model has demonstrated a good accuracy of around 90\%, which is only a 2\% accuracy drop in comparison to its full-precision equivalent. At the same time, a 71-fold reduction in overall energy cost was achieved through model compression. 
 
We keep full-precision inputs and weights in the first and last convolutional layer of the model, whereas the other convolutional layers are binarized. Each convolution layer is followed by a Batch Normalization layer and ReLU activation. Then, the network applies average pooling across the entire image, generating N predictions for each class. Finally, we append a Fully Connected Layer as the output layer of the model, which will be used for CL training. We use the Adam optimizer with an initial learning rate of $10^{-4}$, which is reduced by a factor of ten if training shows no improvement for 10 consecutive epochs. As a loss function, we use categorical cross entropy. The model is trained using TensorFlow (TF) 2.15.

\begin{figure}[t]
    \centering
    \includegraphics[width=.95\columnwidth]{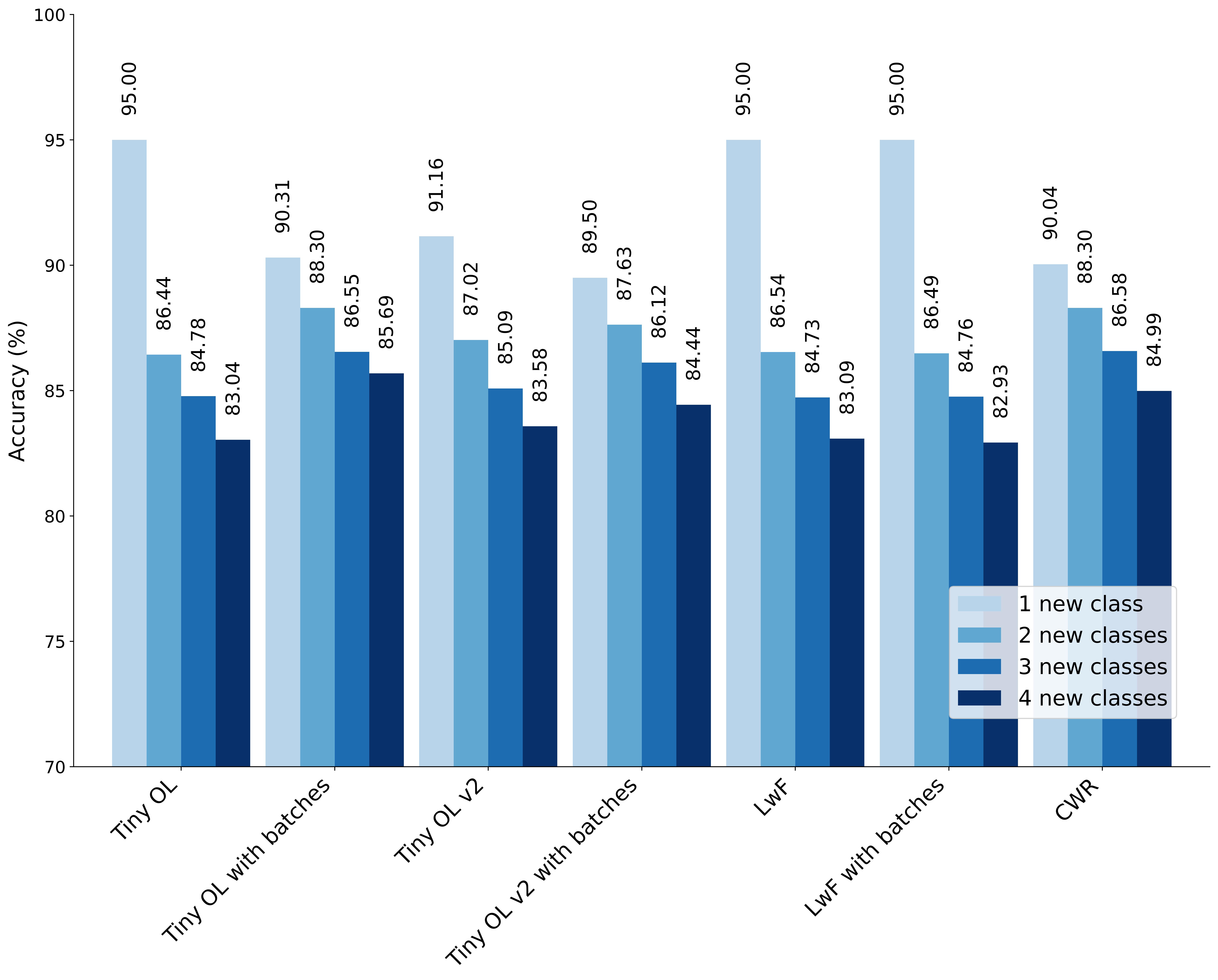}
    \caption{Accuracy of all classes across the seven CL algorithms and different number of new classes.}
    \label{fig_total}
\end{figure}

\subsection{Continual Learning}

After pre-training the BNN model, CL algorithms are applied to adjust the weights and bias of the last Fully Connected layer using backpropagation. During CL, the other layers of the KWS model are frozen. 

Figure \ref{fig_backpropa} describes the CL process. In this step, the model is trained in real-time with a stream of data and labels, representing new samples of already known classes, as well as samples of the new classes in a random order. The weights and biases of the last layer are updated after each new sample is received.  Seven different CL algorithms are used, namely TinyOL, TinyOL with batches, TinyOLv2, Tiny OL v2 with batches, LwF, LwF with batches, and CWR. For algorithms that use batches, we set the batch size to 32. The learning rate is set to 0.05. 

To assess the dependency on the number of new classes, we evaluate the performance of the algorithms with scenarios of 1-4 new classes. We evaluate all potential class combinations, i.e., four combinations for one new class, six combinations for two new classes, and three combinations for three new classes.

\section{Results and Discussion}
\label{s4}
\subsection{Performance of CL algorithms using BNN model}

\begin{table*}[]
\centering
        \caption{Comparison of FLOPs in the backpropagation step of the different CL algorithms}
\begin{tabular}{llcccc}
\toprule
\multicolumn{1}{c}{\multirow{2}{*}{Method}} & \multicolumn{1}{c}{\multirow{2}{*}{Equation}}  & \multicolumn{4}{c}{New classes} \\ \cline{3-6} 
\multicolumn{1}{c}{} & \multicolumn{1}{c}{}                    & 1   & 2   & 3   & 4   \\ \midrule
TinyOL              & $2MN+M+3N$                              & 363 & 390 & 417 & 445 \\
TinyOL w batches                           & $2N+2MN+(3MN+3N+4)/ \text{batch\_size}$        & 354    & 381    & 408   & 436   \\
TinyOL v2           & $2MN+2M+3N$                             & 375 & 402 & 429 & 456 \\
TinyOL v2 with batches                     & $2MN+M+2N+(MM+M+4+3MN+3N)/ \text{batch\_size}$ & 371    & 398    & 425   & 452   \\
LwF                  & $3MN+M+7N+1$                            & 572 & 615 & 658 & 701 \\
LwF with batches               & $3MN+M+7N+1+(MN+N)/ \text{batch\_size}$ & 577 & 620 & 664 & 707 \\
CWR                  & $2MN+3N+M+(5MN+10N)/\text{batch\_size}$ & 391 & 421 & 449 & 479 \\ \bottomrule

\end{tabular}
\label{tab:methods_comparison}
\end{table*}

As our dataset is balanced, we use accuracy as the evaluation metric. Fig.~\ref{fig_12} shows the accuracy for classification of the initial 12 classes, whereas Fig.~\ref{fig_total} shows the accuracy for the entire test set. This enables to determine how effectively the proposed methods retain previously learned knowledge while adapting to newly introduced keywords. The accuracy value represents the average across all possible combinations of newly added keywords for cases where multiple combinations are possible. 

Fig. \ref{fig_12} and \ref{fig_total} demonstrate that the accuracy declines for all of the CL algorithms as more new keywords are introduced. However, it is clear from Fig. \ref{fig_12} that all methods can successfully maintain a high accuracy on the 12 initial classes, even with the addition of four new classes. In this worst-case scenario, the best performing algorithm is TinyOL with batches, with an accuracy of 88.3\%, whereas the worst performing algorithm is LwF with batches, with 86.4\%. This demonstrates only minor differences between algorithms and a performance reduction of approximately 3-5\% with respect to the pre-trained model.

In contrast, for a single new keyword, the differences between algorithms is more prominent. TinyOL, LwF, and LwF with batches achieve the highest accuracies, exceeding 95\%, which even surpasses the accuracy of the initial pre-trained model. The worst performing algorithm in this scenario is CWR with an accuracy of 90.3\%, which is a 1\% reduction with respect to the pre-trained model.

When taking all classes into account (Fig. \ref{fig_total}), the overall trend remains the same. TinyOL with batches and CWR remain to be the best-performing algorithms with four new keywords, whereas TinyOL, LwF and LwF with batches perform best with only one new keyword. Overall, the accuracy is in most cases lower than that for the initial 12 classes, demonstrating that the new classes are not integrated into the model as well as the initial classes. However, the accuracy generally remains high with a worst-case accuracy of 82.9\%.

Fig. \ref{fig_12} and \ref{fig_total} also show that algorithms based on individual samples can achieve higher accuracies in scenarios with one new class, but quickly decrease in performance when more new classes are added. In contrast, batch-based algorithms are less sensitive to the class number.

\subsection{CL algorithm sensitivity to data volume} 

The results presented in Figs. \ref{fig_12} and \ref{fig_total} represent a best-case scenario for CL, as they use all available CL samples in the dataset. However, in practical applications, the number of CL training samples may be limited or should be constrained to achieve resource efficient training. The sensitivity of different algorithms to the number of training samples is thus an important parameter.

To evaluate this sensitivity, we conducted experiments using varying numbers of training samples in the CL phase. The number of training samples was increased exponentially from 64 to 16,384 samples. The experiments were conducted on the worst-case scenario, i.e., adding four new classes. 

\begin{figure}[t]
      \includegraphics[width=1\columnwidth]{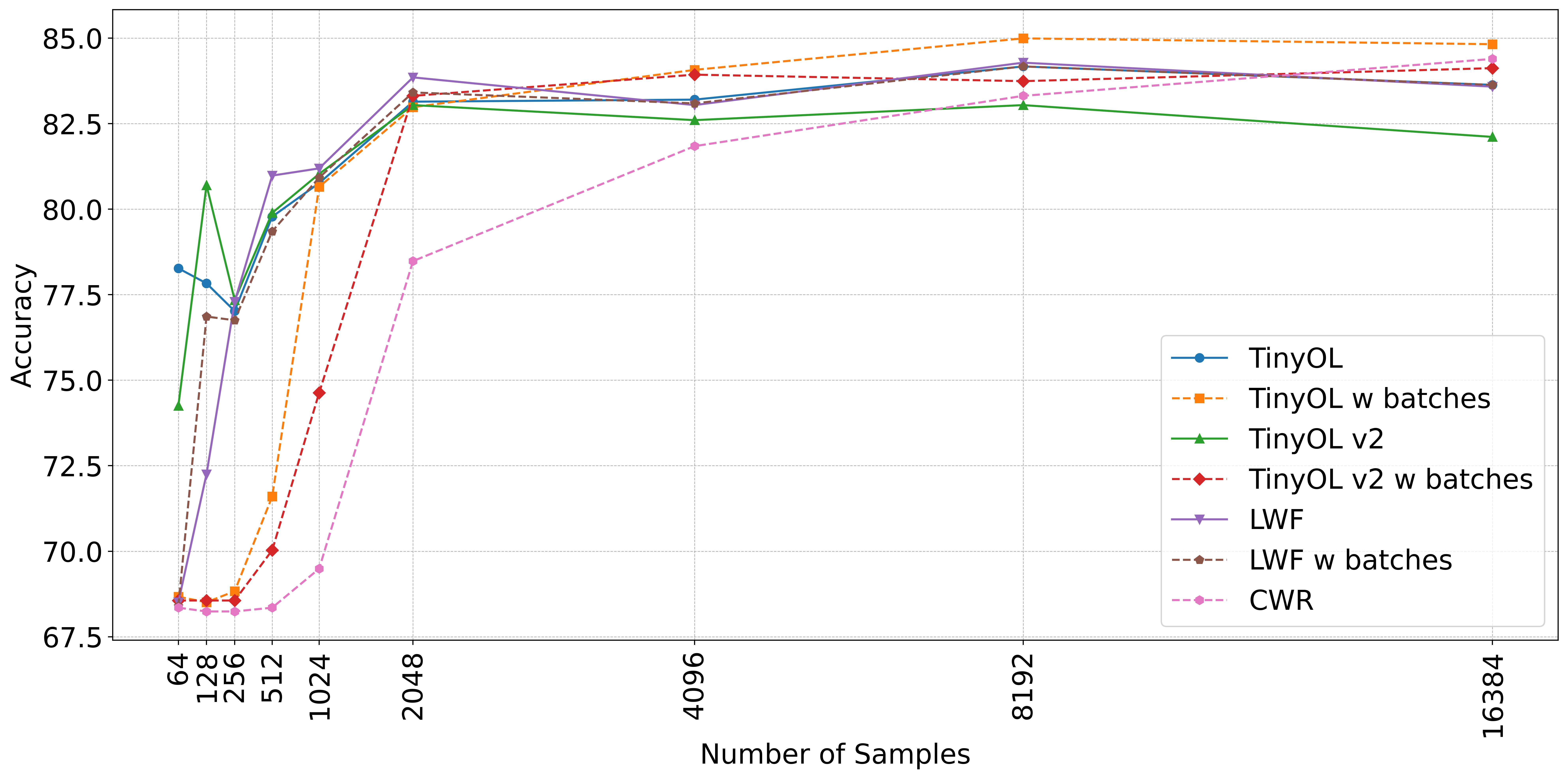}
      \caption{Dependency of the accuracy of the seven CL algorithms on the number of samples used in the CL step.}
      \label{fig_total_dif}
\end{figure}

The results, shown in Fig. \ref{fig_total_dif}, indicate that TinyOL, TinyOL v2, and LwF (with and without batches) achieve stable accuracies at approximately 2048 samples, whereas batch-based methods (excluding LwF with batches) require at least . This suggests that batch-based algorithms demand more data in practical applications. TinyOL and TinyOL v2 consistently achieve higher accuracies with limited data. CWR performs worst with few training samples, but improves significantly when the number of training samples increases.

\subsection{Computational complexity}

To assess the computational complexity of the algorithms, we compute the number of FLOPs involved in the backpropagation step of the CL algorithms for each new sample. The equation of how to calculate the number of FLOPs has been derived in \cite{avi2022incremental} and are listed in Table~\ref{tab:methods_comparison} together with the result for each algorithm. Herein, M represents the number of classes used to train the BNN model, and N denotes the total number of classes (16) in the CL strategy.

The results show that the two LwF variants have the highest computational complexity with 701 and 707 FLOPs, respectively. This is a 55\% increase as compared to the other algorithms, which lie between 436 and 479 FLOPs. The reason for this is that LwF has to compute the loss on the copy layer and the training layer.

Nonetheless, in comparison to the total number of FLOPs for the forward pass of the model, which is approximately 291 MFLOPs, the computational complexity of the backpropagation is negligible.


\section{Conclusion}

\label{s6}

In this study, we evaluated a CL approach for KWS on resource-constrained systems, combining resource-efficient CL algorithms with a BNN model. This approach exploits the benefits of binarized models in terms of computational and memory efficiency, while integrating techniques that facilitate the seamless addition of new keywords over time. We evaluated and compared seven state-of-the-art CL algorithms, measuring their performance not only on newly added keywords but also on the initial ones. 


From the results, several conclusions can be made. First, all evaluated CL algorithms can potentially be used for the proposed BNN-based CL approach, demonstrating the ability to learn new keywords, while maintaining accurate classification capabilities of the existing ones. Although the model accuracy decreases with the number of new classes, accuracies above 82\% were observed even under worst-case conditions. Second, the selection of the most suitable algorithm depends on the number of new classes. While TinyOL and LwF performed best with one additional keyword, TinyOL v2 and CWR demonstrated a more consistent performance and provided the best accuracies for larger number of new classes. Third, batch-based algorithms require a larger number of training samples to perform well, which can limit their usability in real-world applications with limited data or resource constraints. Fourth, computational complexity can be neglected as an evaluation criteria. Although LwF requires a larger amount of FLOPs in the backpropagation, the effect is negligible in comparison to the computational demand in the forward pass, which is the same for all algorithms. 

Further research may focus on the following aspects. A more detailed investigation on the minimum size of the CL dataset may provide additional knowledge on effective training with less data. Here, specifically effects of batch sizes and the utilization of the same data in multiple epochs might be explored. Moreover, deployment of the proposed approach on resource-constrained MCUs should be evaluated.


\section*{Acknowledgement}
The authors would like to acknowledge the financial support of this research by the Knowledge Foundation under grant numbers 2018017 (NIIT) and 20230028-H-01 (IRS TransTech).

\balance
\bibliographystyle{./IEEEtran}
\bibliography{references}

\end{document}